\documentclass[10pt,twocolumn,letterpaper]{article}
\pdfoutput=1

\usepackage{iccv}
\usepackage{times}
\usepackage{graphicx}
\usepackage{caption}
\usepackage{subcaption}
\usepackage{xcolor}
\usepackage{epsfig}
\usepackage{ dsfont }
\usepackage{amsmath}
\usepackage{amssymb}
\usepackage{booktabs}

\usepackage[pagebackref=true,breaklinks=true,letterpaper=true,colorlinks,bookmarks=false]{hyperref}

\iccvfinalcopy 

\def\iccvPaperID{6603} 

\ificcvfinal\fi
\begin{document}

\title{PCRNet: Point Cloud Registration Network using PointNet Encoding}

\author{\text{Vinit Sarode}$^\text{1}$\thanks{equal contribution}\\
\and
\text{Xueqian Li}$^{1*}$\\
\and
\text{Hunter Goforth}$^\text{3}$\\
\and
\text{Yasuhiro Aoki}$^\text{2}$\\
\and
\text{Rangaprasad Arun Srivatsan}$^\text{4}$\\
\and
\text{Simon Lucey}$^\text{1,3}$\\
\and
\text{Howie Choset}$^\text{1}$\\
\and
$^\text{1}$Carnegie Mellon University
\and
$^\text{2}$Fujitsu Laboratories Ltd.
\and
$^\text{3}$Argo AI.
\and
$^\text{4}$Apple.\\
\and
{\tt\small \{vsarode, xueqianl\}@andrew.cmu.edu}
\and
{\tt\small aoki-yasuhiro@fujitsu.com}
\and
{\tt\small \{hgoforth, slucey, choset\}@cs.cmu.edu}
\and
{\tt\small aruns@apple.com}
}

\maketitle

\begin{abstract}
  PointNet has recently emerged as a popular representation for unstructured point cloud data, allowing application of deep learning to tasks such as object detection, segmentation and shape completion. However, recent works in literature have shown the sensitivity of the PointNet representation to pose misalignment. This paper presents a novel framework that uses the PointNet representation to align point clouds and perform registration for applications such as tracking, 3D reconstruction and pose estimation. We develop a framework that compares PointNet features of template and source point clouds to find the transformation that aligns them accurately. Depending on the prior information about the shape of the object formed by the point clouds, our framework can produce approaches that are shape specific or general to unseen shapes. The shape specific approach uses a Siamese architecture with fully connected (FC) layers  and is robust to noise and initial misalignment in data. We perform extensive simulation and real-world experiments to validate the efficacy of our approach and compare the performance with state-of-art approaches. Code is available at \href{https://github.com/vinits5/pcrnet.git}{https://github.com/vinits5/pcrnet}
\end{abstract}

\begin{figure}[t!]
    \centering
    \includegraphics[width=\columnwidth]{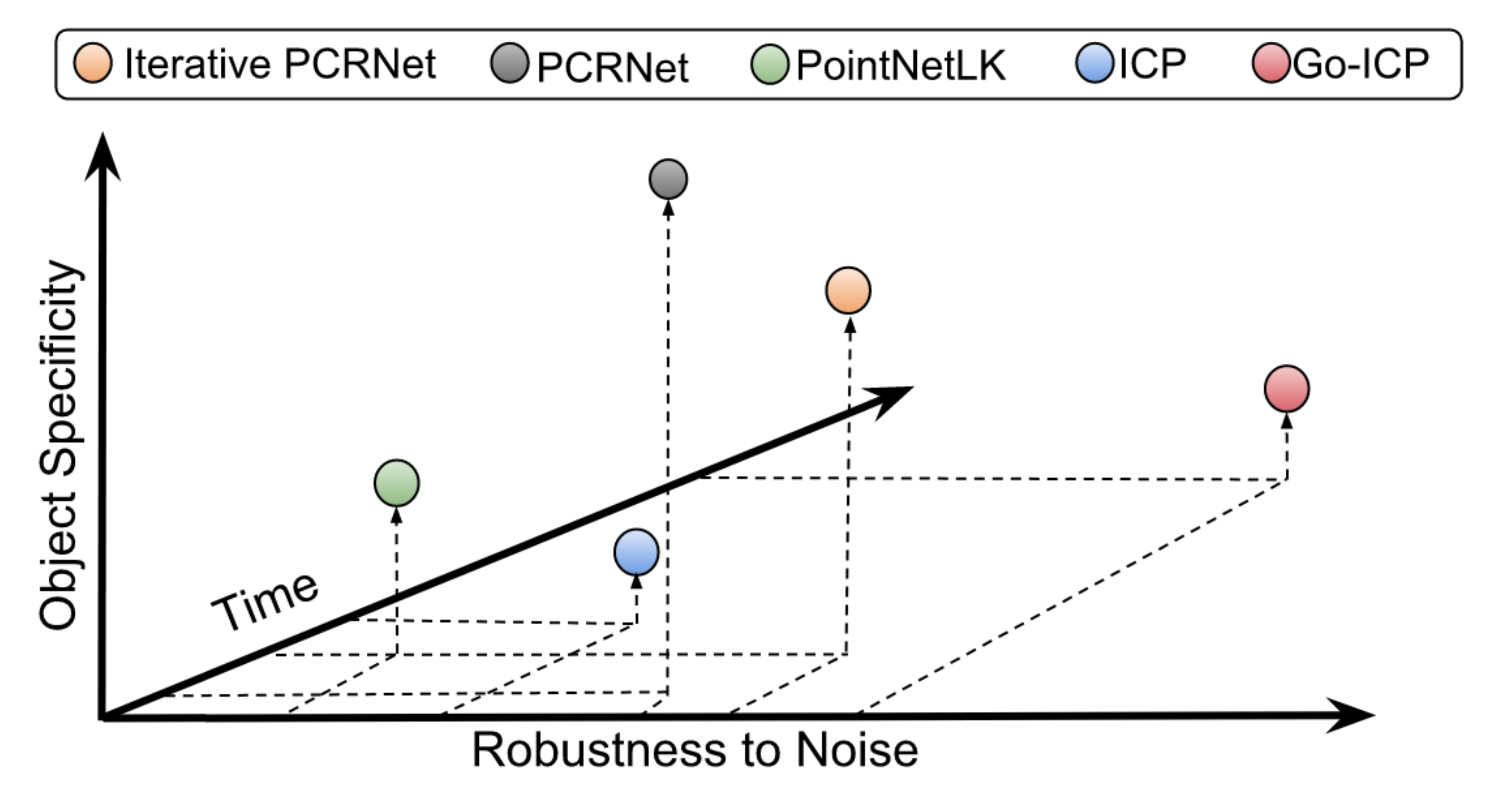}
    \caption{Comparison of different registration methods based on their robustness to noise and computation time with respect to object specificity. The iterative version of PCRNet exploits object specificity to produce accurate results. The PCRNet without iterations is computationally faster but compromises a little on accuracy. PointNetLK~\cite{aoki2019pointnetlk} exhibits good generalizability, but is not robust to noise. ICP~\cite{INTRO:ICP} is object-shape agnostic and slow for large point clouds, while Go-ICP~\cite{RW:GO_ICP} is computationally expensive.}
    \label{fig:firstfig}
\end{figure}

\section{Introduction}
3D point clouds are ubiquitous today, thanks to the development of lidar, stereo cameras and structured light sensors. As a result there has been a growing interest in developing algorithms for performing classification, segmentation, tracking, mapping, etc. directly using point clouds. However, the inherent lack of structure presents difficulties in using point clouds directly in deep learning architectures. Recent developments such as PointNet~\cite{qi2017pointnet} and its variants~\cite{qi2017pointnet++} have been instrumental in overcoming some of these difficulties, resulting in state-of-the-art methods for object detection and segmentation tasks~\cite{RW:frustum,RW:Wentao}. 

Prior works~\cite{RW:Wentao, aoki2019pointnetlk} have observed that robust performance of PointNet requires minimal misalignment of the point clouds with respect to a canonical coordinate frame. While this is present in synthetic datasets such as \emph{ModelNet40}~\cite{wu20153d}, real world data is seldom aligned to some canonical coordinate frame. Inspired by recent works on iterative transformer network (IT-Net)~\cite{RW:Wentao} and PointNetLK~\cite{aoki2019pointnetlk}, this work introduces point cloud registration network (PCRNet), a framework for estimating the misalignment between two point clouds using PointNet as an encoding function. It is worth noting that our approach can directly process point clouds for the task of registration, without the need for hand crafted features~\cite{rusu2009fast,gelfand2005robust}, voxelization~\cite{maturana2015voxnet,Pointregnet} or mesh generation~\cite{wang2018dynamic}. Depending on the prior knowledge of the shape formed by the point clouds, presence of noise, and computational requirements, our framework provides a well-suited approach for each scenario. Our framework also provides additional context for PointNetLK (see Fig.~\ref{fig:firstfig}) within a family of PointNet-based registration algorithms.

Our approach uses PointNet in a Siamese architecture to encode the shape information of a template and a source point cloud as feature vectors, and estimates the pose that aligns these two features using data driven techniques. Using shape-specific prior information in the training phase allows us to be robust to noise in the data, compared to shape agnostic methods such as  iterative closest point (ICP)~\cite{INTRO:ICP} and its variants~\cite{rusinkiewicz2001efficient}. Furthermore, we find that the PointNetLK approach, which uses classical alignment techniques such as Lucas-Kanade (LK) algorithm~\cite{lucas1981iterative,baker2004lucas} for aligning the PointNet features, produces good generalizability to shapes unseen in training but is not robust to noise. Unlike conventional registration approaches such as ICP, our approach does not require costly closest point correspondence computations, resulting in improved computational efficiency and robustness to noise. Further, the approach is fully differentiable which allows for easy integration with other deep networks and can be run directly on GPU without need for any CPU computations.

In summary, our contributions are (1) presenting two novel point cloud alignment algorithms which utilize a PointNet representation for effective registration and (2) a thorough experimental validation of these two approaches including comparison against PointNetLK, ICP, and Go-ICP, on both simulated and real-world data 

\begin{figure*}[t!]
    \centering
    \includegraphics[width=0.9\linewidth]{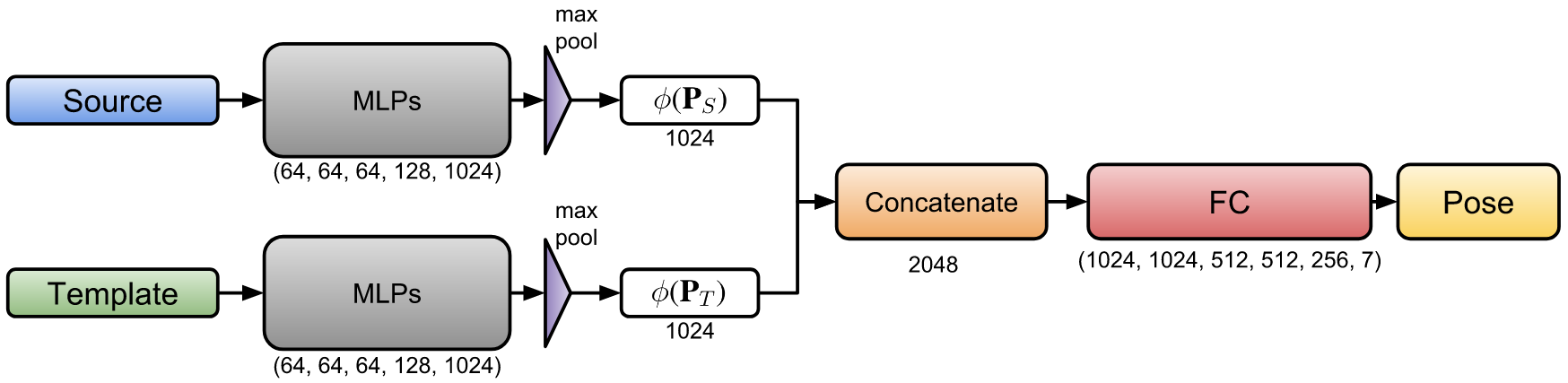}
    \caption{PCRNet Architecture: The model consists of five MLPs having size (64, 64, 64, 128, 1024). The source and template point clouds are sent as input through a twin set of MLPs, arranged in a Siamese architecture. Using a max-pooling function, we obtain global features. Weights are shared between MLPs. These features are concatenated and provided as an input to five fully connected layers 1024, 1024, 512, 512, 256, and an output layer of size 7. The first three output values represent the translation and the last four after normalization represent the rotation quaternion.
}
    \label{fig:one_shot_registration}
\end{figure*}
\section{Related Work}
\paragraph{Classical registration.}
Iterative Closest Point (ICP)~\cite{INTRO:ICP} remains one of the most popular techniques for point cloud registration, as it is straightforward to implement and produces adequate results in many scenarios. Extensions of ICP have added features such as increased computational efficiency~\cite{RW:multi, RW:Gaussion,arun2018probabilistic} or improved accuracy~\cite{RW:GO_ICP}. However, nearly all ICP variants rely on explicit computation of closest points correspondences, a process which scales poorly with the number of points. Additionally, ICP is not differentiable (due to the requirement to find discrete point correspondences) and thus cannot be integrated into end-to-end deep learning pipelines, inhibiting the ability to apply learned descriptors for alignment.

\textit{Interest point} methods compute and compare local descriptors to estimate alignment~\cite{gelfand2005robust,glover2012,guo20143d}. Interest point methods have the advantage of being computationally favorable, however, their use is often limited to point cloud data having identifiable and unique features which are persistent between point clouds that are being registered~\cite{makadia2006fully,ovsjanikov2010one, rusu2009fast}.

\textit{Globally optimal} methods seek to find optimal solutions which cannot reliably be found with iterative techniques such as ICP~\cite{horowitz2014convex, izatt2017, maron2016point}. A representative example which we use as a baseline is Go-ICP~\cite{RW:GO_ICP}, a technique using branch-and-bound optimization. These techniques are characterized by extended computation times, which largely precludes their use in applications requiring real-time speed.

\paragraph{PointNet.}
PointNet~\cite{qi2017pointnet} is the first deep neural network which processes point clouds directly, as opposed to alternative representations such as 2D image projections of objects~\cite{RW:PoseCNN, RW:Semantic,RW:CAD}, voxel representations~\cite{maturana2015voxnet,wu20153d, zhou2018voxelnet} or graph representations~\cite{wang2018dynamic}. Within larger network architectures, PointNet has proven to be useful for tasks including classification, semantic segmentation, object detection~\cite{RW:frustum}, and completion of partial point clouds~\cite{yuan2018pcn}. An extension to PointNet for estimating local feature descriptors is described in~\cite{qi2017pointnet++}. Wentao~\textit{et al.} introduced iterative transformer network (IT-Net)~\cite{RW:Wentao} which uses PointNet to estimate a canonical orientation of point clouds to increase classification and segmentation accuracy. Global descriptors from PointNet are used in~\cite{angelina2018pointnetvlad} for place recognition from 3D data. The loss function used in deep networks for point cloud processing is an important consideration, which we discuss more in Section \ref{section:method}. Earth Mover Distance (EMD) and Chamfer Distance (CD) are introduced in~\cite{fan2017point}, while in~\cite{aoki2019pointnetlk} a Frobenius norm of a difference between estimated and ground truth transformation matrices is used.

\paragraph{Learned registration.}
Discriminative optimization~\cite{vongkulbhisal2017discriminative} and the recent inverse composition discriminative optimization~\cite{vongkulbhisal2018inverse} combine hand-crafted feature vectors and learned map sets for the task of point cloud registration. The shortcoming of these approaches is a quadratic complexity in the number of points, and a lack of generalization due to the feature vector and registration maps both being learned. Deep auto-encoders are used to extract local descriptors for registration of large outdoor point clouds in~\cite{elbaz20173d}. In~\cite{yew20183dfeat}, a network is designed which learns both interest point detection and descriptor computation, for a descriptor-matching registration approach. Wang~\textit{et al.} perform convolution operations on the edges that connect neighboring point pairs, by using a local neighborhood graph~\cite{wang2018dynamic}. PointNetLK~\cite{aoki2019pointnetlk}, which performs registration of arbitrary point clouds by minimizing the distance between the fixed-length, global descriptors produced by PointNet, is the most closely related to our work and serves as a baseline.

\section{Method} \label{section:method}
Point clouds are highly unstructured with ambiguities in the order permutations. While performing classification using PointNet, a symmetric pooling function such as max pool is used to afford invariance to input permutation. The output vector of the symmetry function is referred to as a global feature vector. We will denote the template point cloud \textbf{P}$_T$ and source \textbf{P}$_S$, and the PointNet function $\phi$. Since the global feature vectors contain the information about the geometry as well as the orientation of the point clouds, the transformation between two point clouds can be obtained by comparing the feature vectors. In other words, we calculate the rigid-body transformation $\textbf{T}\in SE(3)$, that minimizes the difference between $\phi$(\textbf{P}$_S$) and $\phi$(\textbf{P}$_T$). 

\subsection{PCRNet}
\label{sec:one-shot registration network}
This section introduces the PCRNet architecture. A block diagram of the architecture is shown in Fig.~\ref{fig:one_shot_registration}. The point cloud data obtained from a sensor is referred to as the \textit{source} and the point cloud corresponding to the known model of the object to be registered is referred to as the \textit{template}.
The model consists of five multi-layered perceptrons (MLPs) similar to the PointNet architecture having size 64, 64, 64, 128, 1024. The MLPs are arranged similar to a Siamese architecture~\cite{held2016learning}.
Both source \textbf{P}$_S$ and template \textbf{P}$_T$ are given as input to the MLPs which are arranged in Siamese architecture and symmetric max-pooling function is used to find the global feature vectors $\phi$(\textbf{P}$_S$) and $\phi$(\textbf{P}$_T$). Weights are shared between MLPs used for source and template.

The global features are concatenated and given as an input to a number of fully connected layers. In this work, we choose five fully connected layers, as they seemed to be sufficient enough for robust performance. We tried using lesser number of FC layers, but the performance of the network was poor.

The FC layers shown by the red block in Fig.~\ref{fig:one_shot_registration} has five hidden layers, 1024, 1024, 512, 512, 256, and an output layer of size 7 whose parameters will represent the estimated transformation \textbf{T}. The first three of the output values we use to represent the translation vector $\mathbf{t}\in \mathds{R}^3$ and last four represents the rotation quaternion $\mathbf{q}\in\mathds{R}^4$, $\mathbf{q}^T\mathbf{q}=1$. In this way, the transformation \textbf{T} which aligns $\phi$(\textbf{P}$_S$) and $\phi$(\textbf{P}$_T$) is estimated with a single forward pass, or single-shot, through the network. The single-shot design lends itself particularly well to high-speed applications, which will be discussed further in Section \ref{sec:results}.

Note that if we were to replace the FC layers in the network with a traditional alignment algorithm such as the Lucas-Kanade~\cite{lucas1981iterative, baker2004lucas}, the resulting implementation would be similar to the PointNetLK~\cite{aoki2019pointnetlk}.

\subsection{Iterative PCRNet}
\begin{figure*}[t]
    \centering
    \includegraphics[width=0.9\linewidth]{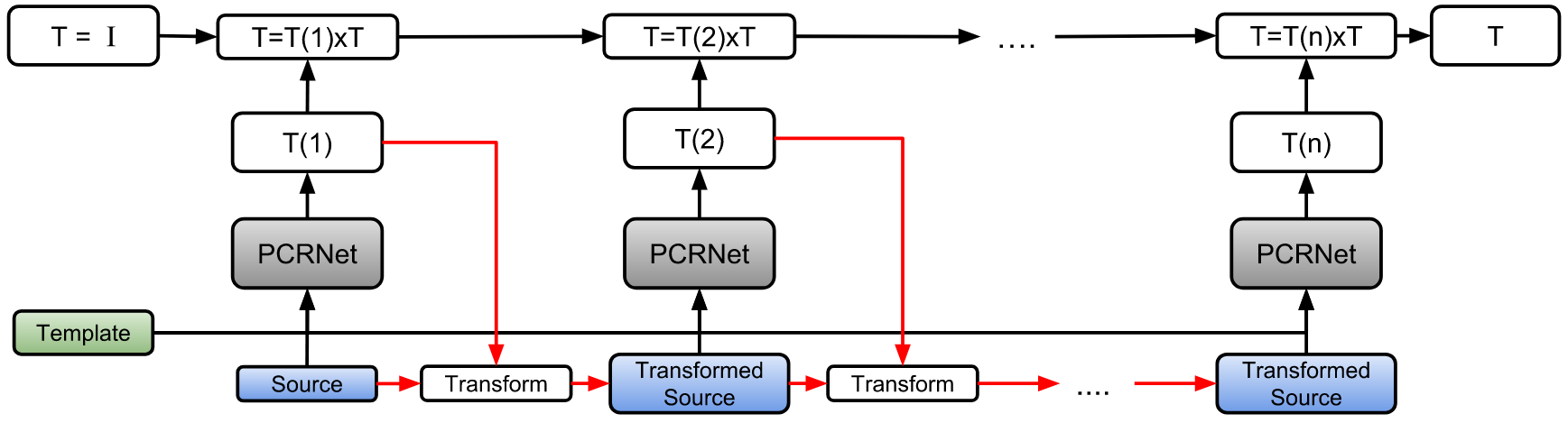}
    \caption{Iterative PCRNet Architecture: The iterative PCRNet uses a modified form of PCRNet described in Fig.~\ref{fig:one_shot_registration} and iteratively improves the estimate of PCRNet. In the first iteration, the source and template point clouds are given to PCRNet which predicts an initial misalignment $\textbf{T}(1)$. The source point cloud is transformed using $\textbf{T}(1)$ and the original template are given as input to the PCRNet, in the next iteration. After performing $n$ iterations, we combined the poses from each iteration to find the overall transformation between the original source and template.
}
    \label{fig:iterative_network}
\end{figure*}
In this section, we present a network with an iterative scheme similar to ICP and Lucas-Kanade  for image alignment as shown in Fig. \ref{fig:iterative_network}. We retain the structure but modify the number of layers from the single-shot PCRNet. For the iterative implementation, the fully connected layers have three hidden layers with size 1024, 512, 256, and an output layer of size seven. Also, there is an additional dropout layer before the output layer, to avoid overfitting. We empirically observe that introducing iterations, allows us to use lesser number of hidden layers compared to PCRNet, and yet obtain robust performance.

In the first iteration, original source and template point clouds are given to PCRNet which predicts an initial misalignment $\textbf{T}(1)$ between them. For the next iteration, $\textbf{T}(1)$ is applied to the source point cloud and then the transformed source and the original template point clouds are given as input to the PCRNet. After performing $n$ iterations, we find the overall transformation between the original source and template point clouds by combining all the poses in each iteration:
\begin{align}
\label{eq:Test}
 \textbf{T} = \textbf{T}(n)\times\textbf{T}(n-1)\times \cdots \times\textbf{T}(1).
 \end{align}

\subsection{Loss Function} The aim of the loss function used to train registration networks should be minimization of distance between the corresponding points in source and template point cloud. This distance can be computed using Earth Mover Distance (EMD) function,
\begin{align}
    \text{EMD}(\textbf{P}_S^{\text{est}}, \textbf{P}_T) = \min_{\psi:\textbf{P}_S^{\text{est}}\rightarrow \textbf{P}_T} \frac{1}{|\textbf{P}_S^{\text{est}}|} \sum_{x\in \textbf{P}_S^{\text{est}}} \|x-\psi(x)\|_2,
\end{align}
where $\textbf{P}_T$ is the template point cloud and  $\textbf{P}_S^{\text{est}}$ is the source point cloud $\textbf{P}_S$, transformed by the estimated transformation $\textbf{T}$ from Eq.~\ref{eq:Test}. This function finds a bijection $\psi$ and minimizes the distance between corresponding points based on $\psi$. While there are many other choices for loss function including Frobenius norm~\cite{aoki2019pointnetlk}, and PoseLoss~\cite{RW:PoseCNN}, we find EMD loss is most effective for learning on the training data described in Section \ref{sec:results} for both iterative and single-shot PCRNet. 

\subsection{Training}
In this work, we use \emph{ModelNet40} dataset~\cite{wu20153d} to train the network. This dataset contains CAD models of 40 different object categories. We uniformly sample points based on face area and then used farthest point algorithm~\cite{eldar1997farthest} to get a complete point cloud.
We train the networks with three different types of datasets as following -- (1) Multiple categories of objects and multiple models from each category, (2) Multiple models of a specific category, (3) A single model from a specific category. We choose these 3 cases to showcase the performance of the PointNet-based approaches on data with differing levels of object-specificity.

We train the iterative PCRNet with 8 iterations during training, observing that more than 8 produced little improvement to results. In some experiments the training data was corrupted with Gaussian noise, which will be discussed in detail in Sec.~\ref{sec:Gaussian_Noise}. The networks are trained for 300 epochs, using a learning rate of $10^{-3}$ with an exponential decay rate of 0.7 after every $3\times10^{6}$ steps and batch size 32. The network parameters are updated with Adam Optimizer on a single NVIDIA GeForce GTX 1070 GPU and a Intel Core i7 CPU at 4.0GHz.

\begin{figure*}[t!]
    \begin{subfigure}{0.49\textwidth}
        \centering
        \includegraphics[width=0.9\linewidth]{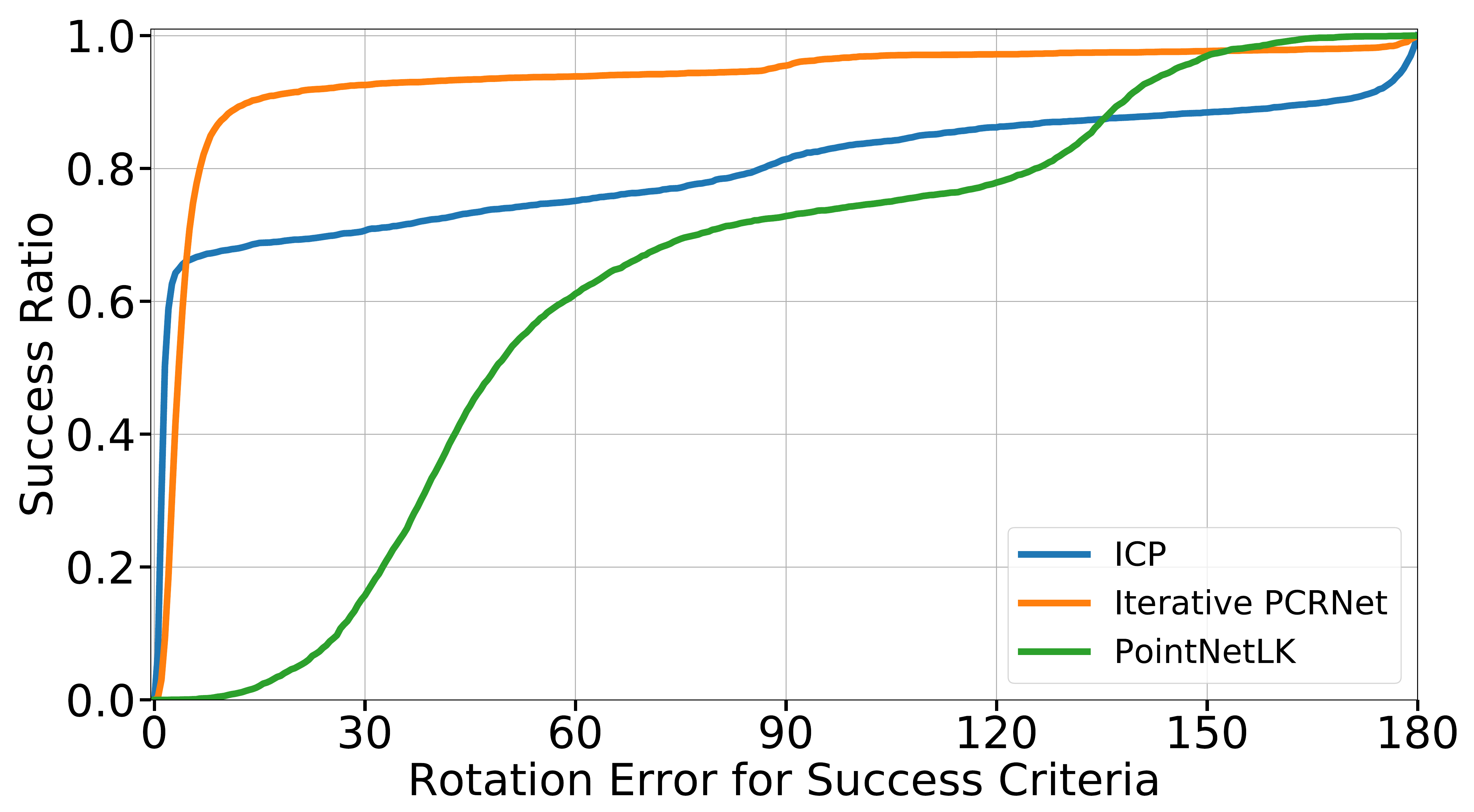}
        \caption{Training and testing: Multiple object categories with noise.}
        \label{fig:multi_catg_noise}
    \end{subfigure}
    ~
    \begin{subfigure}{0.49\textwidth}
        \centering
        \includegraphics[width=0.9\linewidth]{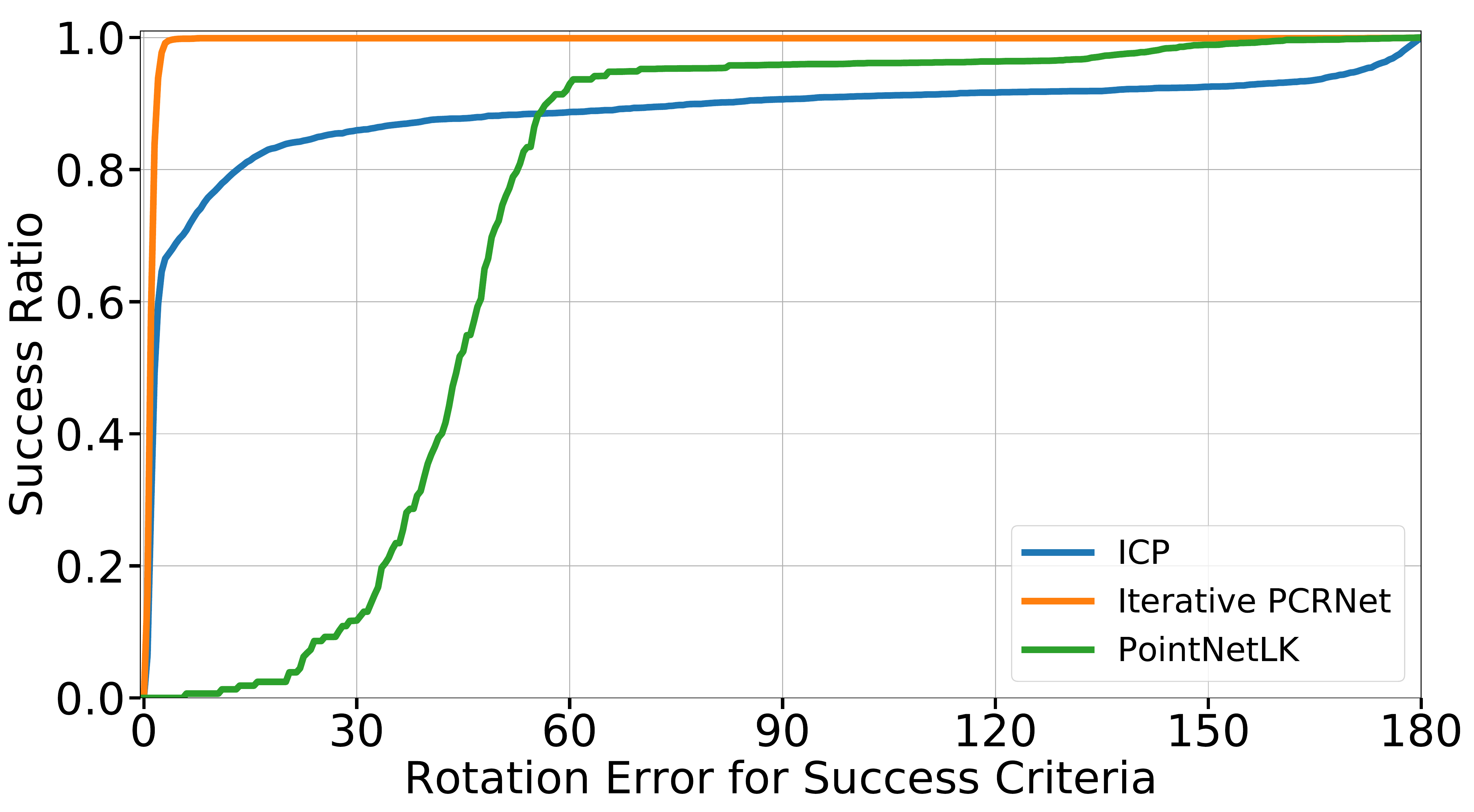}
        \caption{Training and testing: Multiple models of a category with noise}
        \label{fig:one_catg_noise}
    \end{subfigure}
    ~
    \begin{subfigure}{0.49\textwidth}
        \centering
         \includegraphics[width=0.9\linewidth]{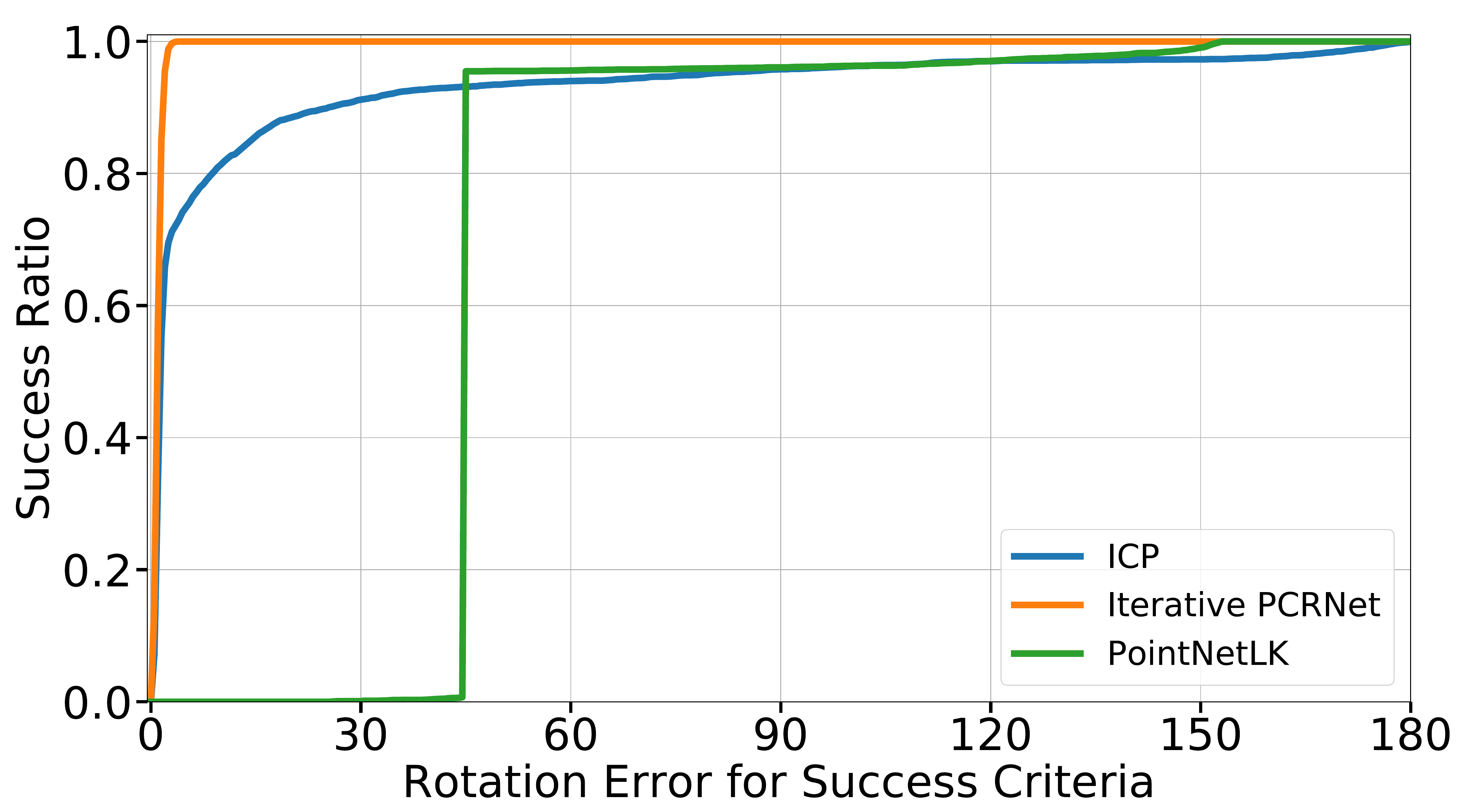}
         \caption{Training and testing: Only one model with noise}
         \label{fig:one_model_noise}
    \end{subfigure}
    ~
    \begin{subfigure}{0.49\textwidth}
        \centering
         \includegraphics[width=0.9\linewidth]{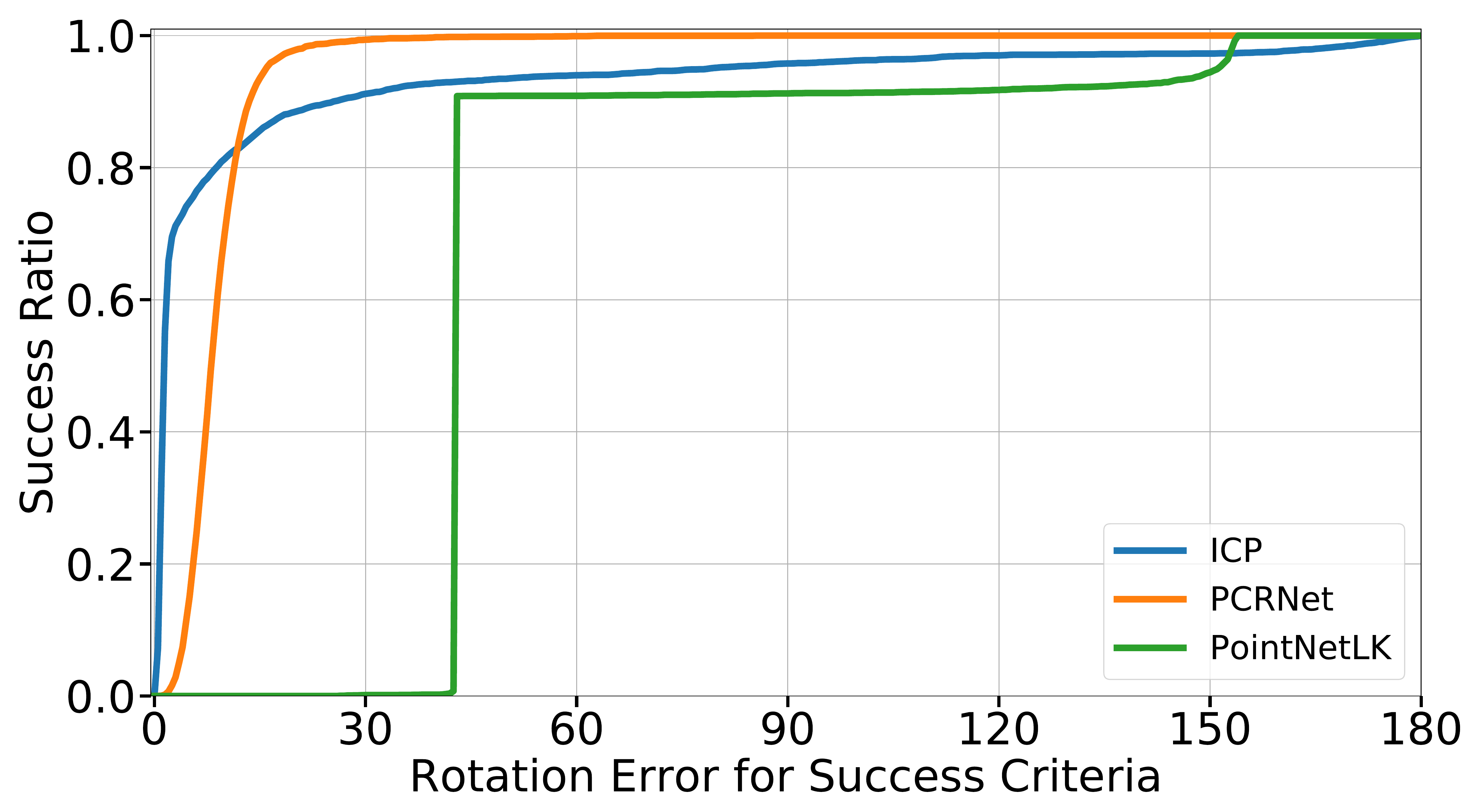}
         \caption{Trained on one model without noise and tested on data with noise}
         \label{fig:one_model_noise_siam}
    \end{subfigure}
    \caption{Results for Section~\ref{sec:Gaussian_Noise}. The $y$-axis is the ratio of experiments that are successful and the $x$-axis shows value of the maximum rotation error that qualifies the estimation to be a success. (a), (b) and (c) shows results for comparisons of iterative PCRNet with ICP and PointNetLK using three different types of datasets. We observe superior performance of iterative PCRNet as our network has more model/category specific information. (d) PCRNet which has not seen noise during training but tested with noisy data also shows good performance and is faster than ICP and PointNetLK. Speed considerations are discussed in Sec.~\ref{sec:speed}.}
    \label{fig:with_noise}
\end{figure*}
\section{Results} \label{sec:results}
In this section, we compare performance of our networks on test data with multiple object categories, a specific object category, a specific object from training dataset and objects unseen in training. We use models from \emph{ModelNet40} dataset~\cite{wu20153d} for the following experiments. Template point clouds are normalized into a unit box and then their mean is shifted to origin. We randomly choose 5070 transformations with Euler angles in the range of $[-45^\circ, 45^\circ]$ and translation values in the range of [-1, 1] units. We apply these rigid transformations on the template point clouds to generate the source point clouds. We allow a maximum of 20 iterations for both iterative PCRNet and PointNetLK while performing tests, while the maximum iterations for ICP was chosen as 100. In addition to maximum iterations, we also use the convergence criteria 
\begin{align*}
\left\| \textbf{T}_{i}\textbf{T}_{i-1}^{-1}-\textbf{I} \right\|_{F} < \epsilon,    
\end{align*}
where $\textbf{T}_{i}, \textbf{T}_{i-1} \in SE(3)$ are the transformations predicted in current and previous iterations, and the value of $\epsilon$ is chosen to be $10^{-7}$.

In order to evaluate the performance of the registration algorithms, a metric we use is area under the curve (AUC). Plots showing success ratio versus success criteria on rotation error (in degrees) are generated for ICP, iterative PCRNet and PointNetLK. Fig. \ref{fig:with_noise} shows examples of these curves. The area below the curves in these plots, divided by 180 to normalize between 0 and 1, is defined as AUC~\footnote{We define success ratio as the number of test cases having rotation error less than success criteria.}.  
AUC expresses a measure of success of registration and so the higher the value of AUC, the better the performance of the network. We measure the misalignment between predicted transformation and ground truth transformation and express it in axis-angle representation and we report the angle as rotation error. As for the translation error, we report the L2 norm of the difference between ground truth and estimated translation vectors. 

\subsection{Generalizability versus specificity}
\label{sec:Generalizability}
In the first experiment, iterative PCRNet and PointNetLK are trained on 20 different object categories from \emph{ModelNet40} with  total of 5070 models. We perform tests using  100 models chosen from 5 object categories which are not in training data (referred as unseen categories) with no noise in point clouds. We ensure that same pair of source and template point clouds are used to test all algorithms, for a fair comparison. 

\begin{figure*}[t!]
    \centering
    \begin{subfigure}{0.3\textwidth}
        \centering
        \includegraphics[width=\linewidth]{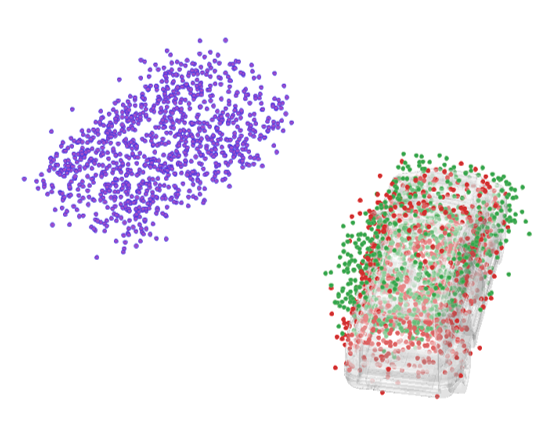}
        \caption{Trained on one car \\ Iterative PCRNet: Rot error = $2.14^\circ$, Trans error = $0.0056$ units.}
        \label{fig:one_model_seen}
    \end{subfigure}
    ~
    \begin{subfigure}{0.3\textwidth}
        \centering
         \includegraphics[width=\linewidth]{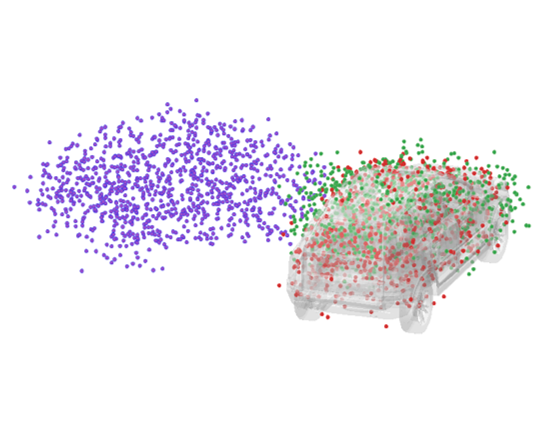}
         \caption{Trained on multiple cars \\ Iterative PCRNet: Rot error = $2.14^\circ$, Trans error = $0.0056$ units.}
         \label{fig:multi_models_seen}
    \end{subfigure}
    ~
    \begin{subfigure}{0.3\textwidth}
        \centering
        \includegraphics[width=\linewidth]{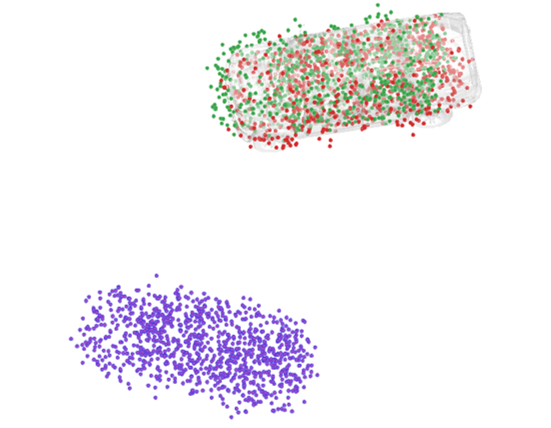}
        \caption{Trained on multiple categories \\ Iterative PCRNet: Rot error = $3.07^\circ$, Trans error = $0.0107$ units.}
        \label{fig:multi_catgs_seen}
    \end{subfigure}
    ~
    \begin{subfigure}{0.3\textwidth}
        \centering
         \includegraphics[width=0.8\linewidth]{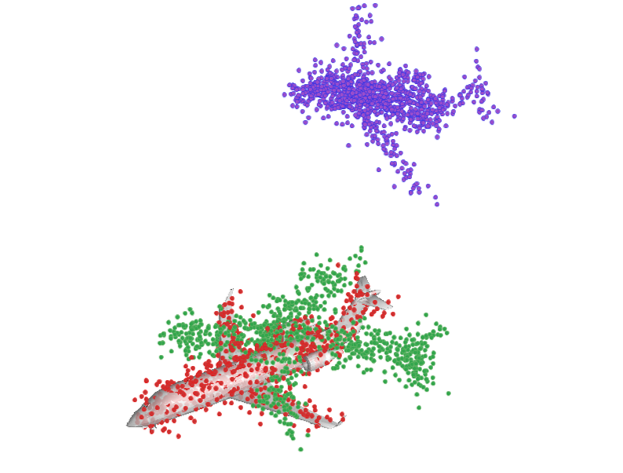}
         \caption{Trained on multiple categories \\ Iterative PCRNet: Rot error = $0.34^\circ$, Trans error = $0.0048$ units. \\ ICP: Rot error = $43.62^\circ$, Trans error = $0.2564$ units.}
         \label{fig:multi_catgs_seen_1}
    \end{subfigure}
    ~
    \begin{subfigure}{0.3\textwidth}
        \centering
         \includegraphics[width=0.8\linewidth]{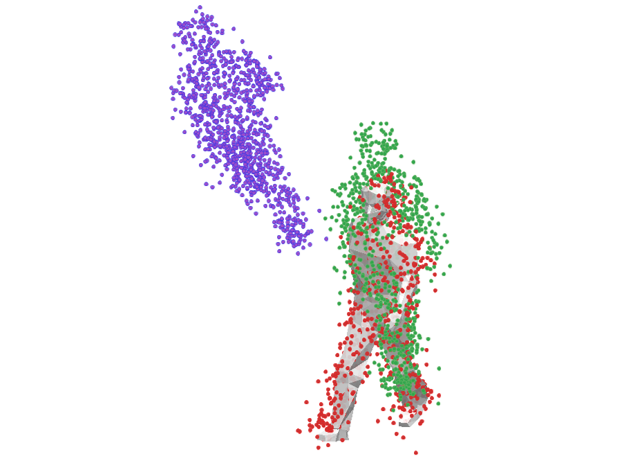}
         \caption{Trained on multiple categories \\ Iterative PCRNet: Rot error = $5.55^\circ$, Trans error = $0.0042$ units. \\ ICP: Rot error = $45.15^\circ$, Trans error = $0.1767$ units.}
         \label{fig:multi_catgs_seen_2}
    \end{subfigure}
    ~
    \begin{subfigure}{0.3\textwidth}
        \centering
         \includegraphics[width=0.8\linewidth]{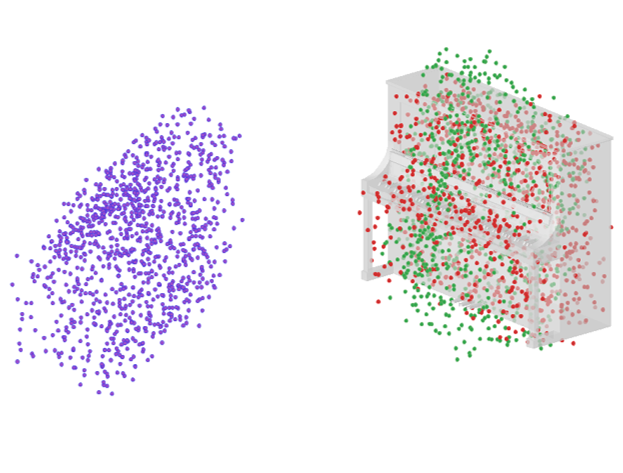}
         \caption{Trained on multiple categories \\ Iterative PCRNet: Rot error = $5.96^\circ$, Trans error = $0.0035$ units. \\ ICP: Rot error = $75.02^\circ$, Trans error = $0.0683$ units.}
         \label{fig:multi_catgs_unseen_1}
    \end{subfigure}
    \caption{Qualitative results for Section \ref{sec:Gaussian_Noise}. For each example, template is shown by a grey rendered CAD model, purple points show initial position of source and red points show converged results of iterative PCRNet trained on data with noise and green points show results of ICP. (a), (b), (c), (d) and (e) show the results for objects from seen categories, while (f) shows results of unseen category. }
    \label{fig:example_registration}
\end{figure*}

We trained iterative PCRNet and PointNetLK using multiple object categories and tested them using object categories which are not in training data. There was no noise in source data during training and testing for this experiment. With these tests, we found that AUC for ICP is 0.802, for our iterative PCRNet is 0.682 and for PointNetLK it is 0.998.

Upon repeating the experiments by training the networks with objects from the same category as the data being tested on, we observe a massive improvement in the AUC for iterative PCRNet, going from 0.682 to 0.972. The AUC for ICP and PointNetLK were similar to earlier at 0.862 and 0.998 respectively, and the AUC of PCRNet was 0.998.

These results emphasize that the iterative PCRNet and PCRNet, when retrained with object specific information, provide improved registration results compared to ICP as well as the version trained with multiple categories. Their performance is comparable to PointNetLK when trained with object specific information. However, PointNetLK shows better generalization than iterative PCRNet across various object categories and has better performance compared to ICP (as also observed by~\cite{aoki2019pointnetlk}). We attribute this to the inherent limitation of the learning capacity of PCRNet to large shape variations, while PointNetLK only has to learn the PointNet representation rather than the task of alignment. However, in the next set of experiments, we demonstrate the definite advantages of PCRNet over PointNetLK and other baselines, especially in the presence of noisy data.

\setlength{\tabcolsep}{4mm}
\begin{table*}[t!]
\centering
\caption[m1]{Results from Section \ref{sec:speed}. Accuracy and computation time comparisons for registering noisy data. Notice that both PCRNet models achieve nearly the same AUC as Go-ICP while being orders of magnitude faster.}
\label{tb:goicp}
\begin{tabular}{lccccccc}\toprule
                 & \multicolumn{2}{c}{Rot. Error (deg)} & \multicolumn{2}{c}{Trans. Error} & \multicolumn{2}{c}{Time (ms)} & AUC \\
 Algorithm                & Mean         & Std. Dev.       & Mean          & Std. Dev.        & Mean       & Std. Dev.   \\ \midrule
PCRNet           & 8.82         & 4.82            & 0.0077        & 0.0008           & 1.89    & 0.398 & 0.9544    \\
Iterative PCRNet & 1.03         & 2.56            & 0.0085        & 0.0024           & 146      & 30.40   & 0.9943   \\
PointNetLK~\cite{aoki2019pointnetlk}       & 51.80        & 29.63           & 0.8783        & 0.0054           & 234      & 41.60  & 0.7059     \\
ICP~\cite{besl1992method}              & 11.87        & 31.87           & 0.0282        & 0.0392           & 407      & 128.00 & 0.9321       \\
Go-ICP~\cite{RW:GO_ICP}           & 0.45       & 0.19           & 0.0016       & 0.0007          & $2.7\times 10^5$   & $1.5\times 10^5$ & 1.0000 \\ \bottomrule       
\end{tabular}
\end{table*}

\subsection{Gaussian noise}
\begin{figure}[htbp]
    \centering
    \includegraphics[width=0.8\columnwidth]{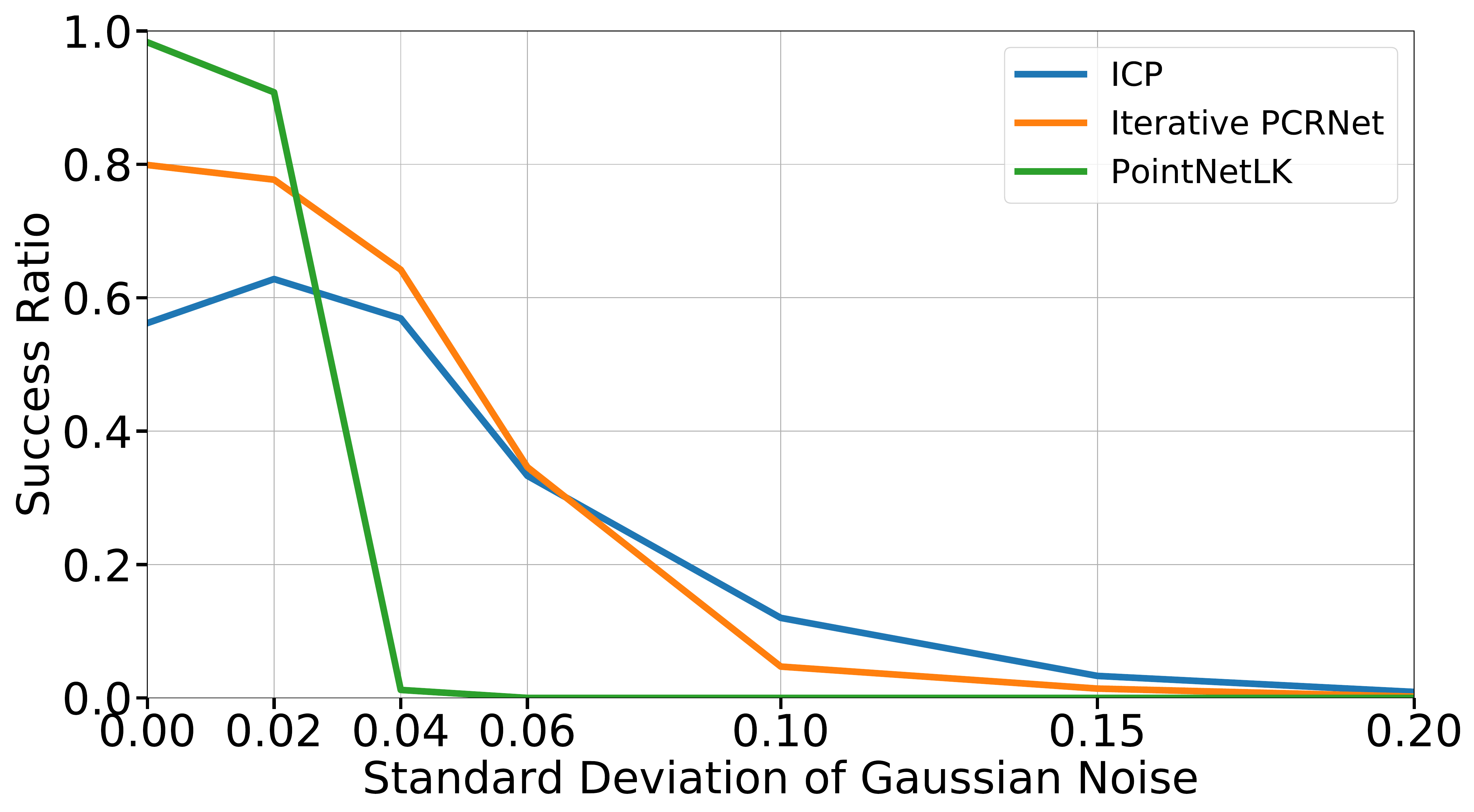}
    \caption{Results for Sec.~\ref{sec:Gaussian_Noise}. Iterative PCRNet and PointNetLK are trained on multiple object categories with Gaussian noise, having maximum value of std. dev. equal to 0.04. The $x$-axis shows different values of standard deviation in noise used in testing. PointNetLK is most accurate in the absence of noise, while iterative PCRNet is robust to noise around the levels that it has observed during training (0.02-0.06).}
    \label{fig:noise_vs_success}
\end{figure}
\label{sec:Gaussian_Noise}
In order to evaluate robustness of our networks to noise, we perform experiments with Gaussian noise in the source points. For our first test, we use dataset as described in Sec.~\ref{sec:Generalizability}. We sample noise from Gaussian distribution for each point in source point cloud with 0 mean and a standard deviation varying in the range of 0 to 0.04 units. For these results, we trained an iterative PCRNet and a PointNetLK which trained with noisy source point clouds using 20 different object categories and a total of 5070 models.

During testing, we compare ICP, PointNetLK and iterative PCRNet with noise in source data for each algorithm. We ensured that the dataset has the same pairs of source and template point clouds for a fair comparison. Fig.~\ref{fig:multi_catg_noise} shows the result. We observe that our iterative PCRNet has higher number of successful test cases with smaller rotation error as compared to ICP and PointNetLK, which shows that our iterative PCRNet is robust to Gaussian noise. It is worth noting that PointNetLK performs the worst and is very sensitive to noisy data. The above test results emphasize that iterative PCRNet works quite well in the presence of noise in the source data, with performance beating all other methods if the object category is known.

For the second test, we used the dataset as described in Sec.~\ref{sec:Generalizability} and added Gaussian noise in source point clouds as described above. We train the networks on a specific object category and test them on the same category using 150 models of cars. Gaussian noise was present during training and testing in source point clouds. The result in Fig.~\ref{fig:one_catg_noise} shows that iterative PCRNet performs the best and has higher number of successful test cases.



\begin{figure}[htbp]
    \centering
    \includegraphics[width=0.9\columnwidth]{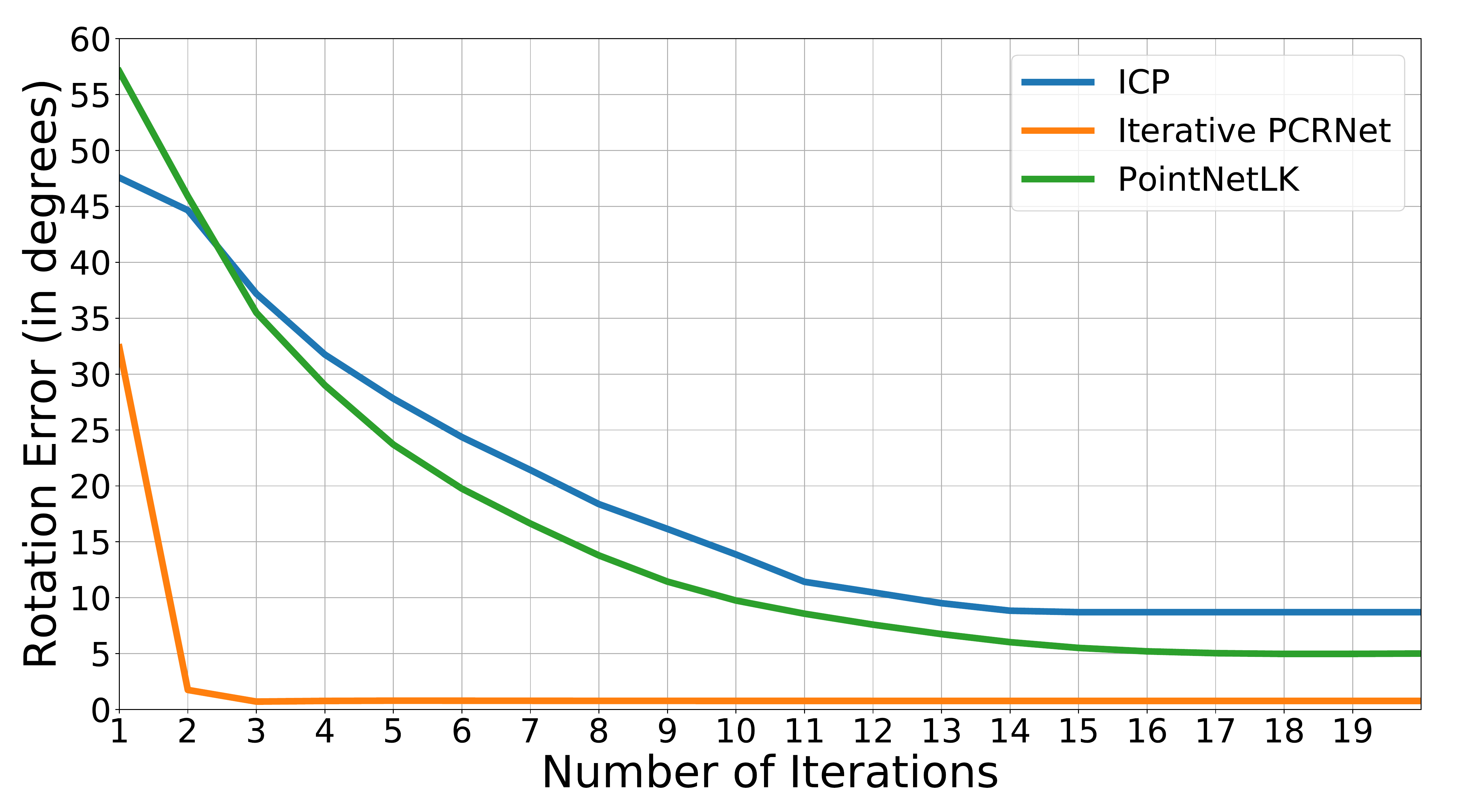}
    \caption{The $y$-axis is rotation error between the predicted and ground truth transformation, and $x$-axis shows the number of iterations performed to find the transformation. Iterative PCRNet shows the ability to align source and template point clouds in fewer iterations. }
    \label{fig:iterations_vs_error}
\end{figure}

We compare the success ratio of networks when training and testing on only one noisy model (see Fig.~\ref{fig:one_model_noise}). Iterative PCRNet once again exhibits a high success ratio, which is better than ICP and PointNetLK. Finally, we compare PCRNet that is trained without noise and tested on noisy data, with ICP and PointNetLK. While not being as good as ICP, our result is still competitive, and performs much better than PointNetLK (See Fig.~\ref{fig:one_model_noise_siam}).

We present qualitative results in Fig.~\ref{fig:example_registration} using iterative PCRNet trained on multiple datasets and testing with noisy data. As expected, the accuracy of iterative PCRNet is highest when trained on the same model that it is being tested on. However, the accuracy drops only a little when trained on multiple models and multiple categories, showing a good generalization as long as there is some representation of the test data in the training. Further the results are accurate also for some unseen categories as shown in Fig.~\ref{fig:example_registration}(c,f), which shows the generalizability of iterative PCRNet.

Fig.~\ref{fig:noise_vs_success} shows success ratio versus change in the amount of noise added to source point clouds during testing. Both iterative PCRNet and PointNetLK are trained on multiple object categories with Gaussian noise having a maximum standard deviation of 0.04. We observe a sudden drop in the PointNetLK performance as the standard deviation for noise increases above 0.02. On the other hand, iterative PCRNet performs best in the neighbourhood of the noise range that it was trained on (0.02-0.06), and produces results comparable to ICP beyond that noise level. This shows that our network is more robust to noise as compared to PointNetLK.

Fig.~\ref{fig:iterations_vs_error} shows the rotation error versus number of iterations in for the different methods. Notice that the iterative PCRNet takes only 3 iterations to get close to convergence, compared to the other methods that take upwards of 15 iterations.
\begin{figure}[htbp]
    \centering
        \includegraphics[width=0.9\linewidth]{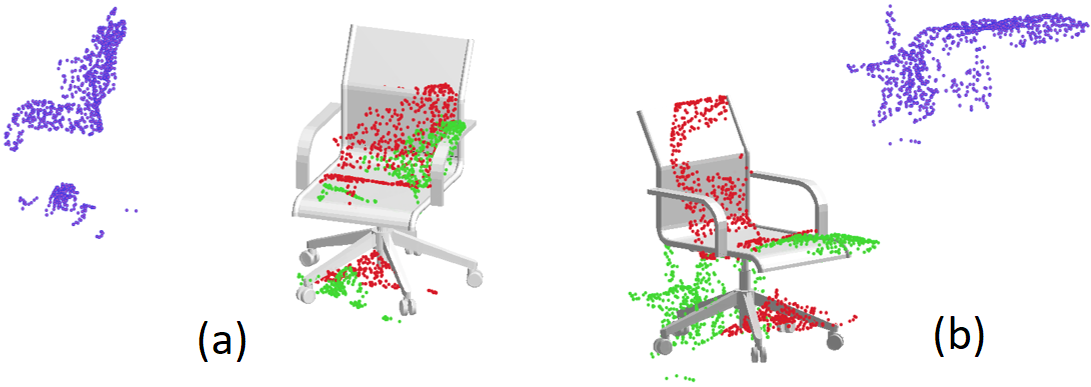}
        \label{fig:room_scene_1}
    \caption{Registration of chair point cloud taken from Stanford \emph{S3DIS} indoor dataset~\cite{armeni_cvpr16}. CAD model shows the template data from \emph{ModelNet40}, purple points is from \emph{S3DIS} dataset, red points represent iterative PCRNet estimates, while the green ones represent ICP estimates.}
    \label{fig:my_label}
\end{figure}

\begin{figure*}[ht]
    \centering
        \includegraphics[width=\linewidth]{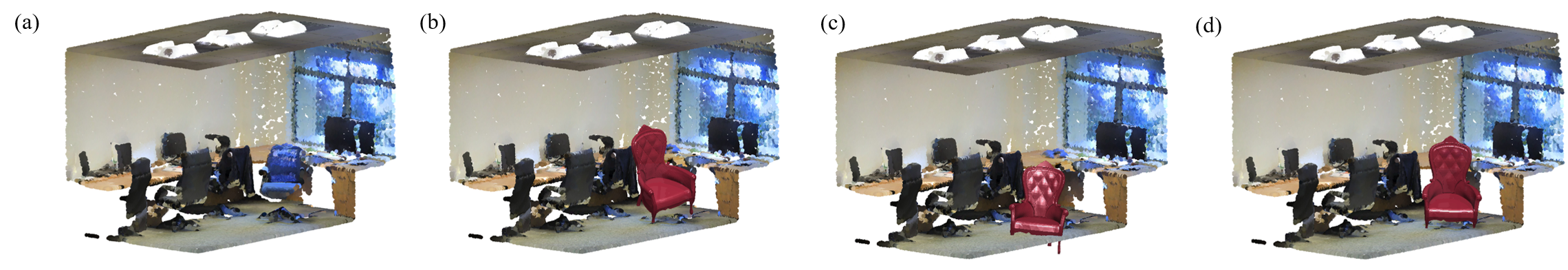}
        \label{fig:room_1}
        \caption{Qualitative results for Section~\ref{sec:replacement}. Replacement of chairs in office scene from Stanford \emph{S3DIS} indoor dataset\cite{armeni_cvpr16}. Red leather chairs shows the replaced chair from \emph{ModelNet40}~\cite{wu20153d} (a) Original scene. Red leather chair replaced by using registration from (b) ICP~\cite{INTRO:ICP}, (c) mixed integer programming~\cite{izatt2017}, and (d) iterative PCRNet. }
    \label{fig:my_label2}
\end{figure*}
\subsection{Computation speed comparisons}
\label{sec:speed}
In this experiment, we use a testing dataset with only one model of car from \emph{ModelNet40} dataset, with Gaussian noise in the source data. We apply 100 randomly chosen transformations with Euler angles in range of $[-45^\circ, 45^\circ]$ and translation values in range of [-1, 1] units. The networks are all trained using multiple models of same category (i.e. car). We compared the performance of iterative PCRNet, PCRNet, PointNetLK, ICP and Go-ICP, as shown in Table~\ref{tb:goicp}. We report the rotation and translation error after registration, computation time, and the AUC.

The results demonstrate that Go-ICP converges to a globally optimal solution with a very small rotation error and translation error, but the time taken is three orders of magnitude more than iterative PCRNet and five orders of magnitude more than PCRNet. The AUC value of Go-ICP is 1, meaning that it has converged in all test cases, while our network has the second best AUC value. This experiment shows how the iterative PCRNet is similar to Go-ICP in terms of accuracy, but computationally much faster, allowing for use in many practical applications. Further, while the PCRNet is less accurate than iterative PCRNet and Go-ICP, the accuracy may be good enough as a pre-aligning step in applications such as object detection and segmentation~\cite{RW:Wentao}.

\section{Model replacement using segmentation}
\label{sec:replacement}
To show qualitative performance on real-world data, we demonstrate the use of our approach to find the pose and modify the models in an indoor point cloud dataset~\cite{armeni_cvpr16}. we perform model replacement in use the semantic segmentation network introduced in PointNet~\cite{qi2017pointnet} to predict labels for each object in the Stanford \emph{S3DIS} indoor dataset~\cite{armeni_cvpr16}. Point cloud corresponding to a chair are selected from the scene and registered to a chair model from \emph{ModelNet40} dataset using iterative PCRNet, which was trained on multiple object categories with noise (see Fig.~\ref{fig:my_label}).

The transformation predicted by iterative PCRNet is then applied to the model chosen from \emph{ModelNet40} and replaced at the place of the original chair as shown in Fig.~\ref{fig:my_label2}. The original scene is shown in Fig.~\ref{fig:my_label2}(a). The blue chair is replaced with a red chair of a different model from \textit{ModelNet40} dataset. Fig.~\ref{fig:my_label2}(b) shows the result from using ICP. Notice how ICP fails to register the chair to the right pose. While we observed in Sec.~\ref{sec:speed} that Go-ICP produces the most accurate results, in this example Go-ICP did not improve upon the result of ICP. Thus we report the result of another global registration method that uses mixed integer programming (MIP)~\cite{izatt2017} (see Fig.~\ref{fig:my_label2}(c)). Note that neither ICP nor MIP produce results that are as accurate as that produced by the iterative PCRNet in Fig.~\ref{fig:my_label2}(d). This is because ICP, Go-ICP and MIP require the template and the source to be of the same object and any variation between objects of the same category can result in poor registration. Our approach however, is robust to changes in shapes within the same category and produces a better result.

\section{Discussions and future work}
This work presents a novel data-driven framework for performing registration of point clouds using the PointNet representation. The framework provides an approach to use data specific information for producing highly accurate registration that is robust to noise and initial misalignment, while being computationally faster than existing methods. The framework can be implemented in an iterative manner to obtain highly accurate estimates comparable to global registration methods. The framework could also be implemented without the iterations, but with  deeper layers to produce two to five orders of magnitude speed improvement compared to popular registration methods. The framework illustrates how data-driven techniques may be used to learn a distribution over appearance variation in point cloud data, including noisy data or category-specificity, and perform better at test time using such a learned prior. Finally, this framework also puts into context other recent PointNet-based registration methods in literature such as the PointNetLK. 

Future work would involve modifying the network to handle partial and occluded point clouds, as well as integration into larger deep neural network systems, for tasks such as multi-object tracking, style transfer, mapping, etc. Future work may explore the limitations of the learning capacity of the fully-connected registration layers to the size of data distribution.


{\small
\bibliographystyle{ieee}
\bibliography{egbib,cvprReferences}
}

\end{document}